\documentclass[lettersize,journal]{IEEEtran}
\pdfoutput=1
\usepackage{amsmath,amsfonts}
\usepackage{algorithmic}
\usepackage{algorithm}
\usepackage{array}
\usepackage[caption=false,font=normalsize,labelfont=sf,textfont=sf]{subfig}
\usepackage[inline]{enumitem}
\usepackage{textcomp}
\usepackage{stfloats}
\usepackage{adjustbox}
\usepackage{multirow}
\usepackage{url}
\usepackage{orcidlink}
\usepackage{authblk}
\usepackage{verbatim}
\usepackage{graphicx}
\usepackage{lipsum}
\usepackage{cite}
\usepackage{hyperref} 
\hypersetup{colorlinks,allcolors=black}
\graphicspath{{Figures/}}
\begin{document}

\title{Controlled Dropout for Uncertainty Estimation}

\author{Mehedi Hasan\orcidlink{0000-0001-7721-0258},
        Abbas Khosravi\orcidlink{0000-0001-6927-0744},~\IEEEmembership{Member,~IEEE,}
        Ibrahim Hossain,~
        Ashikur Rahman,~
        Saeid Nahavandi\orcidlink{0000-0002-0360-5270},~\IEEEmembership{Fellow,~IEEE.}
\thanks{Mehedi Hasan, Abbas Khosravi, Ibrahim Hossain and Saeid Nahavandi are with the Institute for Intelligent Systems Research and Innovation, Deakin University, Geelong, VIC 3220, Australia. E-mail: \{mmhasan, abbas.khosravi, i.hossain, saeid.nahavandi\}@deakin.edu.au}
\thanks{Ashikur Rahman is with the Department of Computer Science and Engineering, Bangladesh University of Engineering and Technology, Dhaka 1000,
Bangladesh. E-mail: ashikur@cse.buet.ac.bd}
}


\IEEEpubid{0000--0000/00\$00.00~\copyright~2021 IEEE}

\maketitle

\begin{abstract}
Uncertainty quantification in a neural network is one of the most discussed topics for safety-critical applications. Though Neural Networks (NNs) have achieved state-of-the-art performance for many applications, they still provide unreliable point predictions, which lack information about uncertainty estimates. Among various methods to enable neural networks to estimate uncertainty, Monte Carlo (MC) dropout has gained much popularity in a short period due to its simplicity. In this study, we present a new version of the traditional dropout layer where we are able to fix the number of dropout configurations. As such, each layer can take and apply the new dropout layer in the MC method to quantify the uncertainty associated with NN predictions. We conduct experiments on both toy and realistic datasets and compare the results with the MC method using the traditional dropout layer. Performance analysis utilizing uncertainty evaluation metrics corroborates that our dropout layer offers better performance in most cases.
\end{abstract}


\section{Introduction}
Due to high-level performance, Deep Neural Networks (DNN)\cite{Goodfellow-et-al-2016} have become the pivotal component both in academia and industry for many applications such as autonomous driving\cite{Bojarski2016EndTE}  or medical diagnostics\cite{6868045}. For these safety-critical applications, any wrong decision might put a human at life-threatening risks. For example, in May 2016, a man died in a car accident cause the perception system of his autonomous car confused the white side of a trailer for a bright sky\cite{NHTSA2017}. The system could alert the passenger by assigning a higher uncertainty value to its prediction. Therefore, we need DNNs to tell us when they are not sure about their predictions.

Researchers have formulated different approaches to estimate predictive uncertainty in NNs for the last few years. One of the classical methods is applying Bayesian inference to NN, where all the network parameters are treated as random variables. Instead of putting single values on the parameters, we assign distribution over them. Posterior distributions are estimated assuming suitable priors on them given the training data. This network is called Bayesian Neural Network (BNN)\cite{10.5555/2986766.2986882}\cite{10.1162/neco.1992.4.3.448}\cite{pmlr-v54-sun17b}\cite{pmlr-v70-louizos17a}\cite{ritter2018a}. Thus, it offers a technique to estimate uncertainty in the model's prediction but with increased computational cost\cite{pmlr-v37-blundell15}. Since the denominator of the Bayes formula to calculate the posterior of the network's parameters become intractable, we cannot estimate the exact posterior. Various approximation methods, such as variational inference (VI)\cite{pmlr-v80-zhang18l}\cite{NIPS2011_7eb3c8be}\cite{pmlr-v48-louizos16}, Markov Chain Monte Carlo\cite{neal2012bayesian}, or Laplace approximation\cite{10.1162/neco.1992.4.3.448}\cite{ritter2018a}  have been proposed to estimate the posterior distribution. The ultimate prediction is derived from the expected prediction of the network marginalized over the posterior distributions of the parameters. 

Researchers have investigated different modeling choices when estimating posterior distribution using the Variational Inference (VI) method. In \cite{pmlr-v48-louizos16}, authors  employed a matrix variate Gaussian on parameter posterior distribution to train BNNs. Multiplicative normalizing flow (MNF) was proposed in \cite{pmlr-v70-louizos17a} to model the approximate posterior as a compound distribution, where a normalizing flow provides the mixing density. In \cite{pmlr-v80-zhang18l}, Zhang et al. proposed to use noisy Kronecker-Factored Approximate Curvature (K-FAC) algorithms. The proposed method applies the approximate natural gradient\cite{10.1162/089976698300017746} based maximization on the VI objective.  

A simple but effective approach, named deep ensemble, has been proposed to provide a high-quality uncertainty estimate by researchers in \cite{10.5555/3295222.3295387}. The authors suggested utilizing an ensemble of neural networks in conjunction with a proper scoring rule. They applied random initialization of parameters and data shuffling for each member of the ensemble. However, the trainable parameters become M times higher, where M is the number of ensemble members.

Another straightforward idea has been presented in \cite{pmlr-v48-gal16} to obtain uncertainty estimates from DNN. The simplicity of the method and its easy and direct integration with the existing DNN framework makes it a popular choice in many cases. In their work, the authors demonstrated that dropout, a simple method used to address the overfitting issue in DNNs\cite{JMLR:v15:srivastava14a}, can be approximated as Bayesian variational inference. This technique has proved to be effective in cases such as increasing visual relocalization accuracy\cite{7487679} and semantic segmentation performance on images\cite{BMVC2017_57}.

\IEEEpubidadjcol
Plausible merits and accompanied costs for using MC dropout in typical NN models have been discussed in \cite{Seoh2020QualitativeAO}. In \cite{mukhoti2020batch}, Monte Carlo batch normalization, a parallel technique to MC dropout, was proposed as an approximate inference technique in NNs. The repeatability of NNs while using Monte Carlo dropout was investigated in \cite{2022arXiv220207562L}. In \cite{9439951}, authors developed an advanced dropout technique, a model-free methodology, it adjusts dropout rates to extend the overfitting reduction capability of DNNs. A continuous relaxation of dropout’s discrete masks was deployed in \cite{10.5555/3294996.3295116}. The proposed technique, named `Concrete Dropout', offers better-calibrated uncertainty estimates. It allows automatic tuning of dropout rates using a principled optimization objective. In \cite{gomez2019learning}, the targeted dropout was suggested to leverage robustness to subsequent pruning when training a large and sparse network using a simple self-reinforcing sparsity criterion. In \cite{shamsi2021improving}, authors combined Expected Calibration Error (ECE) and Predictive Entropy (PE) with cross-entropy to form two new loss functions. The proposed loss functions improve the uncertainty estimate of the MC dropout model. Authors in \cite{NEURIPS2021_5a66b920} suggested a simple consistency training strategy to regularize dropout,
namely R-Drop, to push the output distributions of different dropout sub-models to be consistent with each other. The regularization was achieved by minimizing the bidirectional Kullback–Leibler (KL) divergence between the output distributions of two dropout sub-models. In \cite{theobald:hal-03122764}, a dropout regulation method was proposed to control the dropout rate based on the performance difference between training and validation data. Authors in \cite{Sicking2020WassersteinD} proposed a purely non-parametric technique, `Wasserstein dropout' for regression tasks, and aleatoric uncertainty was captured by utilizing a new objective that minimizes the Wasserstein distance in dropout-based sub-network distributions. We refer interested readers to this review paper\cite{ABDAR2021243} for a comprehensive discussion on existing works.

In most papers related to dropout, authors tried to modulate dropout rates using different techniques. In \cite{NEURIPS2021_5a66b920}, they tried to control the dropout sub-models. To the best of our knowledge, no one has ever tried to control the number of dropout sub-models or dropout configurations of each layer. In this work, we present a variation of the dropout method where we constrain the number of dropout configurations that any layer can follow and investigate the effect of our approach in uncertainty estimation on both toy and realistic classification datasets. We name our method `Controlled dropout'.

The rest of the paper is structured as follows. Section \ref{sec:background} briefly introduces uncertainty in NNs, MC dropout and uncertainty evaluation metrics. Section \ref{sec:Study Proposal} presents the proposed methods and algorithms. Section \ref{sec:Experimental Setup} explains the experimental setup. Section \ref{sec:result} discusses the results. Sections \ref{sec:Conclusion} concludes the paper.

\section{Background}
\label{sec:background}
\subsection{Uncertainty in Neural Network}
At first, it needs to be clarified what types of uncertainty we can face when dealing with an NN. There is an endless debate defining and categorizing uncertainties in scientific modeling. Without going into any of that, we will follow the categorization presented by Der Kiureghian \& Ditlevsen\cite{KIUREGHIAN2009105}. Mainly, we will encounter two types of uncertainty associated with an NN's predictions:
\begin{itemize}
    \item{\textbf{Epistemic Uncertainty:} Also referred to as \textit{model uncertainty}, this type arises when the model is not sure of its knowledge about the distribution of data. It implies that we do not know the appropriate weight values of the network. It may happen because of the lack of data in certain input domains. Theoretically, this uncertainty can be eliminated by observing more data and careful collection of clean data. Following that, it is also called the \textit{reducible uncertainty}.
    }
    \item{\textbf{Aleatoric Uncertainty:} Also called \textit{data uncertainty}, it arises from the inherent stochasticity of the data generation process. Reasons behind this type of uncertainty might be measurement errors during capturing data or any missing elements that should be considered (lack of observation). Based on its dependency on the input data, we can further divide this uncertainty into \textit{homoscedastic} and \textit{heteroscedastic uncertainty}. \textit{Aleatoric Uncertainty} cannot be reduced by collecting more data. Therefore, it is also called the \textit{irreducible uncertainty}.
    }
\end{itemize}
In this work, we will concentrate on epistemic uncertainty as it is directly related to how we develop NNs.
\subsection{Dropout for Uncertainty Estimation}
Consider an NN with an arbitrary number of hidden layers. Let's assume \textbf{u} as the input to any layer and \textbf{y} as the output from that layer. If weights and biases of any layer are expressed by \textbf{W} and \textbf{b}, then the standard feed-forward operation in any layer $l$ of a network is defined by:
\begin{equation}
    \textbf{u}^l=\textbf{W}^l\textbf{y}^{l-1}+\textbf{b}^l
\end{equation}
\begin{equation}
    \textbf{y}^l=f(\textbf{u}^l)
\end{equation}
where, $f(.)$ denotes any non-linear activation function.

In the presence of dropout, the forward equation becomes\cite{JMLR:v15:srivastava14a}:
\begin{equation}\label{eq:bernoulli}
    \textbf{z}^{l-1} \sim Bernoulli(1-p)
\end{equation}
\begin{equation}
    \Tilde{\textbf{y}}^{l-1} = \textbf{y}^{l-1}*\textbf{z}^{l-1}
\end{equation}
\begin{equation}
    \textbf{u}^l=\textbf{W}^l\Tilde{\textbf{y}}^{l-1}+\textbf{b}^l
\end{equation}
\begin{equation}
    \textbf{y}^l=f(\textbf{u}^l)
\end{equation}
where p indicates the dropout rate and * implies the element-wise multiplication. \textbf{z} is a vector where its elements are sampled from an independent Bernoulli distribution with the probability of $1-p$. With the introduction of \textbf{z} into the network, a random sub-network from the whole network is chosen in each forward pass. Usually, the dropout is turned off during the test time. It turns out that dropout is an easy solution to the over-fitting problem of a network\cite{JMLR:v15:srivastava14a}.

In \cite{pmlr-v48-gal16}, Gal et al. conducted a theoretical treatment of a network having dropouts in its layers. It was demonstrated that we could consider such a network as an approximation to a Bayesian network. This enables us to estimate the uncertainty of a model from the traditional neural network approach without much change. The only thing that we need to do is to keep the dropout layers active during test time and draw multiple output samples for each input. This method is named Monte Carlo (MC) dropout. Uncertainty can then be estimated using predictive entropy (PE). PE is calculated using the following formula:
\begin{equation}
PE=-\sum_c\frac{1}{T}\sum_tp(y=c|x,\hat{w}_t)\log{\frac{1}{T}\sum_t p(y=c|x,\hat{w}_t)}
\end{equation}
where c is the output class, T is the number of samples per input and $\hat{w}_t$ is the weight matrix for the network during $t^{th}$ forward pass for any input.
\subsection{Uncertainty Evaluation Metrics}
Normally, model predictions are grouped into two categories: correct and incorrect. For uncertainty estimation, we have another two groups: certain and uncertain predictions\cite{Asgharnezhad2022}. We also need to have some threshold values to categorize predictions into certain and uncertain. Combining these categories at hand, four groups are formed: 
\begin{enumerate*}[label=(\roman*)]
\item True Certain (TC): the prediction is correct and the model is certain about its prediction,\label{item:TC}
\item True Uncertain (TU): the prediction is incorrect and the model is not sure about its prediction,\label{item:TU}
\item False Certain (FC): the model predicts incorrectly with certainty, and\label{item:FC}
\item False Uncertain (FU): the model is uncertain though the prediction is correct.\label{item:FU}
\end{enumerate*}
TU and TC are the ideal cases, and FU might be acceptable based on the situation, while FC is the worst case that one aims to avoid as much as possible.

Confusion matrix is an effective tool to estimate the performance of a model in classification problems where accuracy alone can be a misleading indicator due to the uneven distributions of classes. Performance metrics similar to regular confusion matrix metrics are also defined for uncertainty estimation as below\cite{Asgharnezhad2022}:
\begin{itemize}
\item{Uncertainty Sensitivity ($U_{Sen}$): Sensitivity indicates how much of the positive outputs are correctly labeled as positive. In case of uncertainty estimation, one would like to flag the wrong prediction as uncertain. Following that, $U_{Sen}$ becomes the ratio of incorrect predictions that are uncertain (TU) to all the wrong predictions (TU+FC). This metric holds paramount importance as a higher value of $U_{Sen}$ implies that the model confidently makes very few incorrect predictions:
\begin{equation}
    U_{Sen}=\frac{TU}{TU+FC}
\end{equation}
}
\item{Uncertainty Specificity ($U_{Spec}$): Specificity is also referred to as the true negative rate. $U_{Spec}$ is defined as the ratio of correct predictions that are certain (TC) to all the right predictions (TC+FU). A high value of $U_{Spec}$ implies that the model is certain about most of the correct predictions:
\begin{equation}
    U_{Spec}=\frac{TC}{TC+FU}
\end{equation}
}
\item{Uncertainty Precision ($U_{Prec}$): Precision is also called the positive predictive value. $U_{Prec}$ is defined by the ratio of uncertain and incorrect (TU) predictions to all uncertain predictions (TU+FU). A high value of $U_{Prec}$ indicates that most of the wrong predictions are uncertain:
\begin{equation}
    U_{Prec}=\frac{TU}{TU+FU}
\end{equation}
}
\item{Uncertainty Accuracy ($U_{Acc}$): $U_{Acc}$ is defined as the ratio of the sum of correct certain (TC) and incorrect uncertain (TU) predictions to all predictions made by the model. A high value of $U_{Acc}$ is desirable from a well-trained model. 
\begin{equation}
    U_{Acc}=\frac{TU+TC}{TU+TC+FC+FU}
\end{equation}
} 
\end{itemize}
For an ideal model, these metrics will be close to 1. Higher values in $U_{Prec}$, $U_{Spec}$ and $U_{Sen}$ indicate that a model clearly knows what it does not know.

\section{Study Proposal}
\label{sec:Study Proposal}
\begin{figure}[!t]
\centering
\includegraphics[width=2.5in]{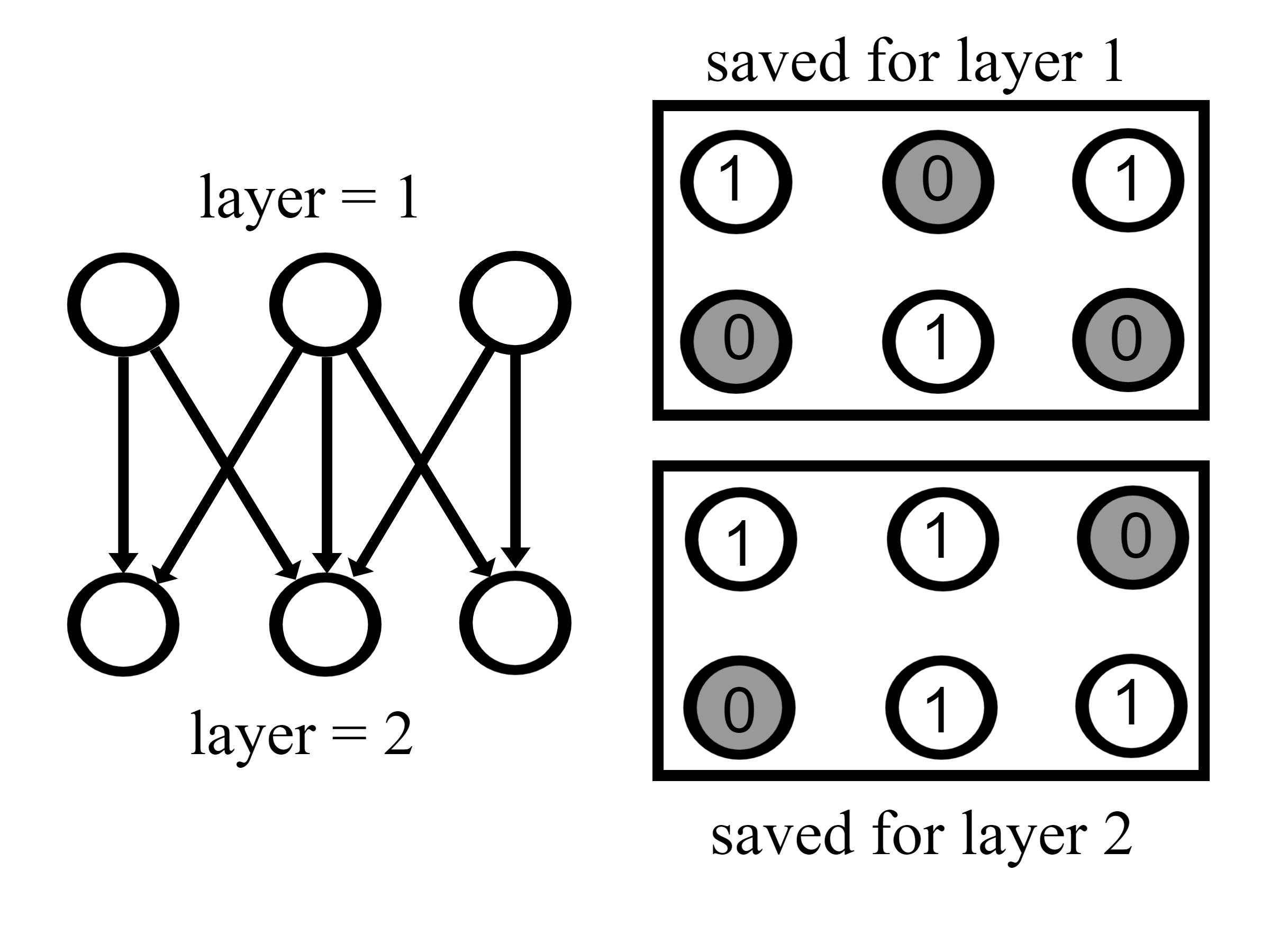}
\caption{2 layer network with saved dropout structures}
\label{fig1}
\end{figure}

Let's assume that an NN consists of two hidden layers, each of which consists of three units (left side of Fig. \ref{fig1}). If we apply the dropout on these two layers, each layer can take seven different configurations, such as one unit `on' or two units `on'. Consequently, 49 different dropout configurations are possible for the two-layer network. The actual number of configurations depends on the dropout probability. As per the current trend, NNs have many layers and units in each layer. As a result, the number of possible configurations with dropout in one or more layers becomes extremely large. For example, if we apply dropout rate of 0.5 to a network with a single hidden layer having 100 units, the number of dropout configurations will be more than $10^{29}$. It is almost impossible to get all the structures during both training and testing time using the MC dropout method for the accurate uncertainty estimation.
\begin{algorithm}[!t]
\caption{Algorithm to create instance of Controlled dropout module}\label{alg:alg1}
\begin{algorithmic}
\STATE \textbf{Input:} $\mathbf{S}$ = size of the dropout layer,\\$\mathbf{N_{sample}}$ = Number of layer samples to be stored,\\$\mathbf{p}$ = dropout rate for the layer
\STATE \textbf{Output:} Instance of CMC dropout layer with stored dropout configurations $\mathbf{C_{sample}}$
\STATE \hspace{0.5cm}\textbf{Initialize} an empty vector $\mathbf{z}$ of length $\mathbf{S}$ and an empty list $\mathbf{C_{sample}}$
\STATE \hspace{0.5cm}\textbf{while} No. of stored vector in $\mathbf{C_{sample}}<\mathbf{N_{sample}}$
\STATE \hspace{1cm}\textbf{do}
\STATE \hspace{1cm}\textbf{Sample} each element of $\mathbf{z}$ from $Bernoulli(1-p)$
\STATE \hspace{1cm}\textbf{Divide} each element of $\mathbf{z}$ by $1-p$
\STATE \hspace{1cm}\textbf{if} $\mathbf{z}$ is unique to all the saved vectors in $\mathbf{C_{sample}}$
\STATE \hspace{1.5cm}\textbf{do}
\STATE \hspace{1.5cm}\textbf{Append} $\mathbf{z}$ to $\mathbf{C_{sample}}$
\STATE \hspace{1cm}\textbf{end if} 
\STATE \hspace{0.5cm}\textbf{end while} 
\end{algorithmic}
\end{algorithm}
In this study, we intend to see the effect of constraining the number of configurations of the dropout layer of an NN on the uncertainty metrics. Let's get back to our first example of 2 layer network. If we constrain the number of dropout configurations of each layer to two (right side of Fig. \ref{fig1}) irrespective of the dropout probability, then the total number of configurations generated from this network will always be four. The actual dropout structures can be produced from a stochastic process implying that the saved configurations may vary from one instance of the network to the other instance. We name this dropout process ``Controlled Dropout''. Although the number of structures may be limited, still for each forward pass and each input, a new realization for binary vector \textbf{z} is sampled from the saved structures instead of using (\ref{eq:bernoulli}) and the same values of \textbf{z} are also used in the backward pass.

We have depicted our approach in Algorithm \ref{alg:alg1} and \ref{alg:alg2}. In  Algorithm \ref{alg:alg1}, we save $\mathbf{N_{sample}}$ unique dropout structures, each of which is generated from a stochastic process. Here, we mention that there could be cases where a particular choice of $p$ and $\mathbf{S}$ does not allow $\mathbf{N_{sample}}$ unique structures of size $\mathbf{S}$ to exist. To tackle such a case, some constraints can be applied to $p$, $\mathbf{S}$ and $\mathbf{N_{sample}}$.

The forward operation is depicted in  Algorithm \ref{alg:alg2} where the obvious difference is observed from the traditional dropout layer. For the normal dropout layer, the vector $\mathbf{z}$ is generated from a Bernoulli distribution during each forward pass. But in our case, we uniformly sample the vector $\mathbf{z}$ from a fixed set of vectors already stored in $\mathbf{C_{sample}}$. One subtle thing to mention here is that rather than passing $\mathbf{C_{sample}}$ as an input to the forward method and sampling the vector $\mathbf{z}$ inside of the forward method, we can take the sample $\mathbf{z}$ when calling the forward function and pass only the sampled $\mathbf{z}$ as a parameter to the function. Other basic operations for both approaches remain the same.

\begin{algorithm}[!t]
\caption{Operations during forward pass}\label{alg:alg2}
\begin{algorithmic}
\STATE \textbf{Input:} $\mathbf{x}$ = input to the dropout layer,\\$\mathbf{C_{sample}}$ = stored dropout configurations
\STATE \textbf{Output:} $\mathbf{y}$ = output of dropout layer
\STATE \hspace{0.5cm}\textbf{if} in training mode \textbf{do}
\STATE \hspace{1.0cm}\textbf{Sample} a vector $\mathbf{z}$ uniformly from $\mathbf{C_{sample}}$
\STATE \hspace{1cm}$\mathbf{y} \gets \mathbf{x}*\mathbf{z}$
\STATE \hspace{0.5cm}\textbf{else} 
\STATE \hspace{1cm}$\mathbf{y} \gets \mathbf{x}$
\STATE \hspace{0.5cm}\textbf{end if} 
\end{algorithmic}
\end{algorithm}

For the rest of the paper, when using our dropout layer with the MC dropout method, we refer to it as the CMC method. When a model uses the CMC method, we refer to it as the CMC model.
\section{Experimental Setup}
\label{sec:Experimental Setup}
We perform experiments on two toy datasets and the MNIST dataset to demonstrate the performance of the CMC method for classification problems.
\subsection{Two Moon and Circle Datasets}
To investigate the performance of our method, we chose the two moon and circle datasets. 10000 samples are generated with a noise level of 0.3 for each dataset. We intentionally keep the noise level bit high to make the investigation fair. The dataset is randomly split into training, validation, and test set having 8000, 1000, and 1000 samples, respectively.

For the NN, an architecture with three hidden layers is selected where each layer consists of 20 units. The dropout is applied to the first two hidden layers. For our CMC model, we randomly generate and store 10 structures for each dropout layer. The stochastic gradient descent optimization is adopted using a learning rate of 0.08. We apply the binary cross-entropy loss as the loss metric. All models are trained for 300 epochs. After each epoch, we evaluate the performance of the model on the validation dataset. The best-performing model in terms of the loss value on the validation set is saved for the final calculation using the test dataset. We apply 100 forward passes to produce 100 output samples per input for uncertainty estimation. All models are deployed, trained, and tested using PyTorch framework\cite{NEURIPS2019_9015}.

\begin{figure*}[!t]
\subfloat[]{\includegraphics[width=0.45\columnwidth]{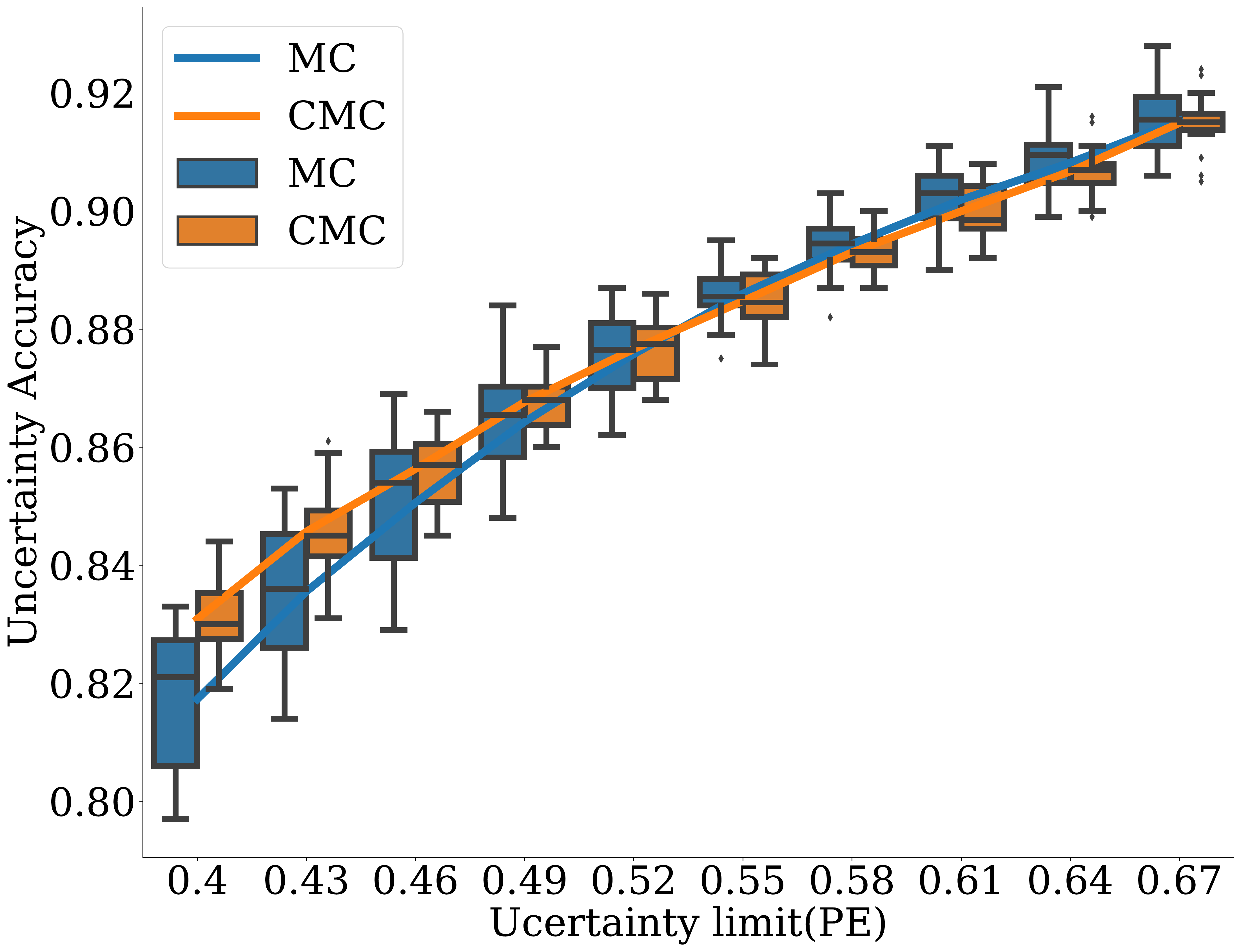}%
\label{moon_acc_uncer}}
\hfil
\subfloat[]{\includegraphics[width=0.45\columnwidth]{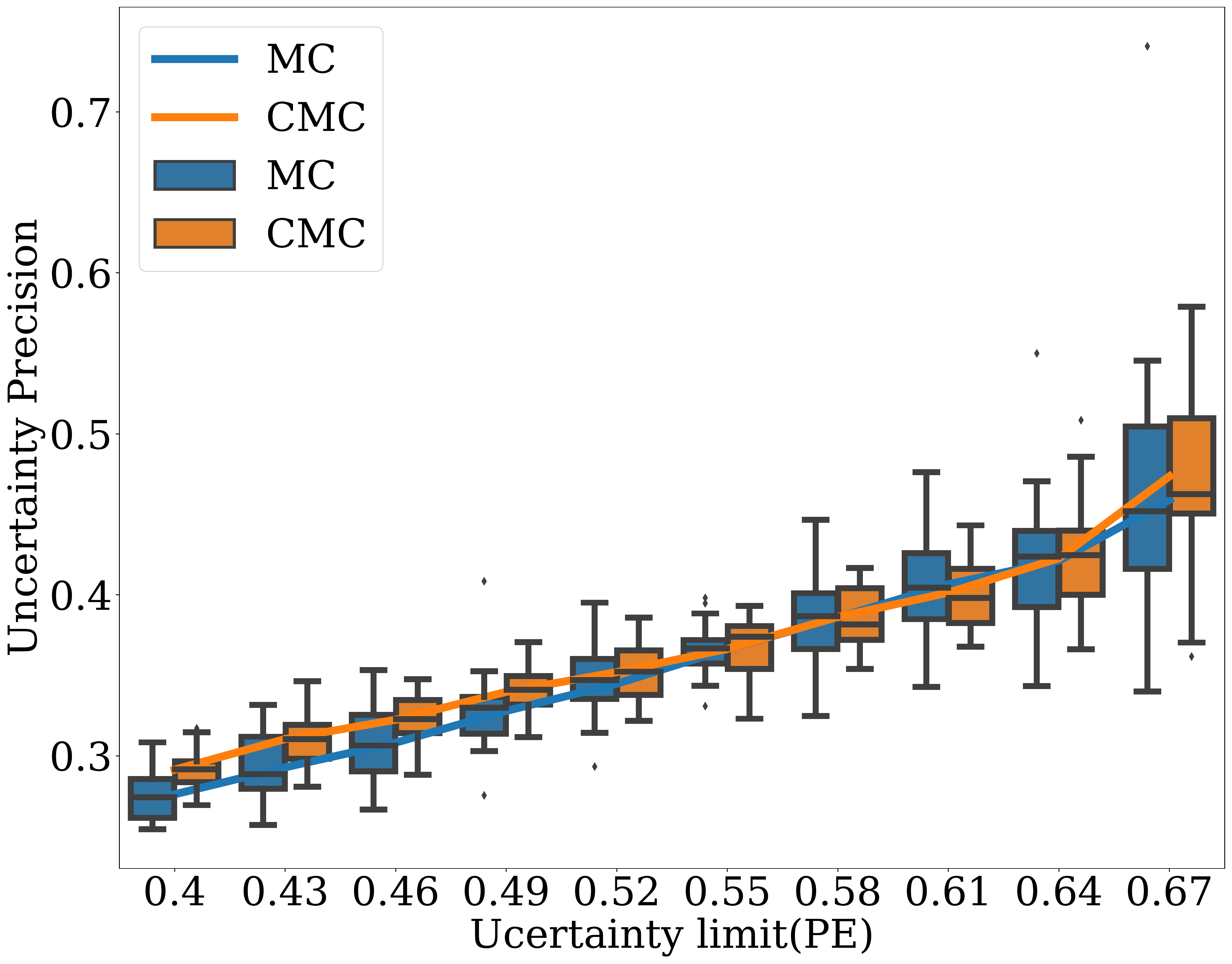}%
\label{moon_prec_uncer}}
\hfil
\subfloat[]{\includegraphics[width=0.45\columnwidth]{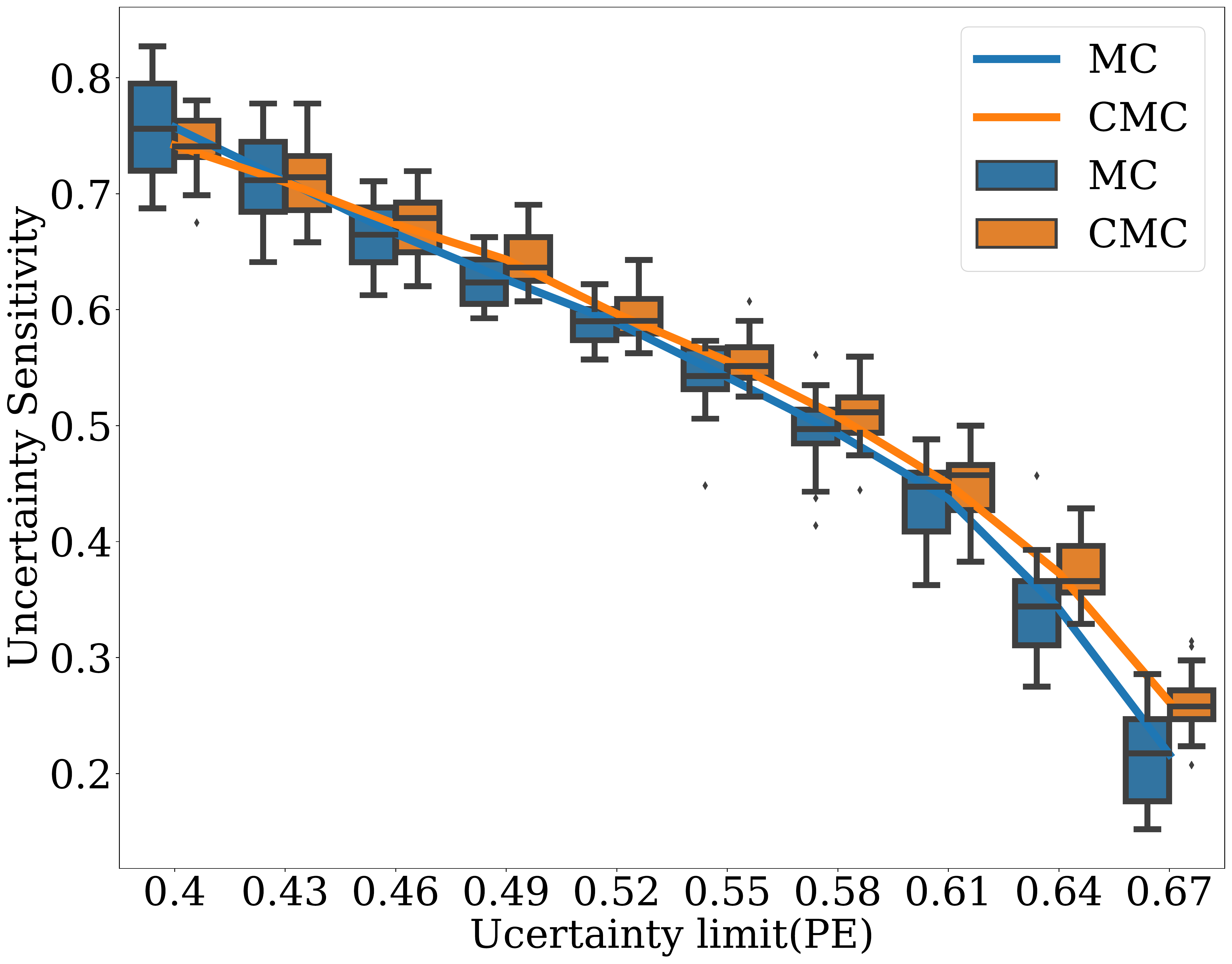}%
\label{moon_sen_uncer}}
\hfil
\subfloat[]{\includegraphics[width=0.45\columnwidth]{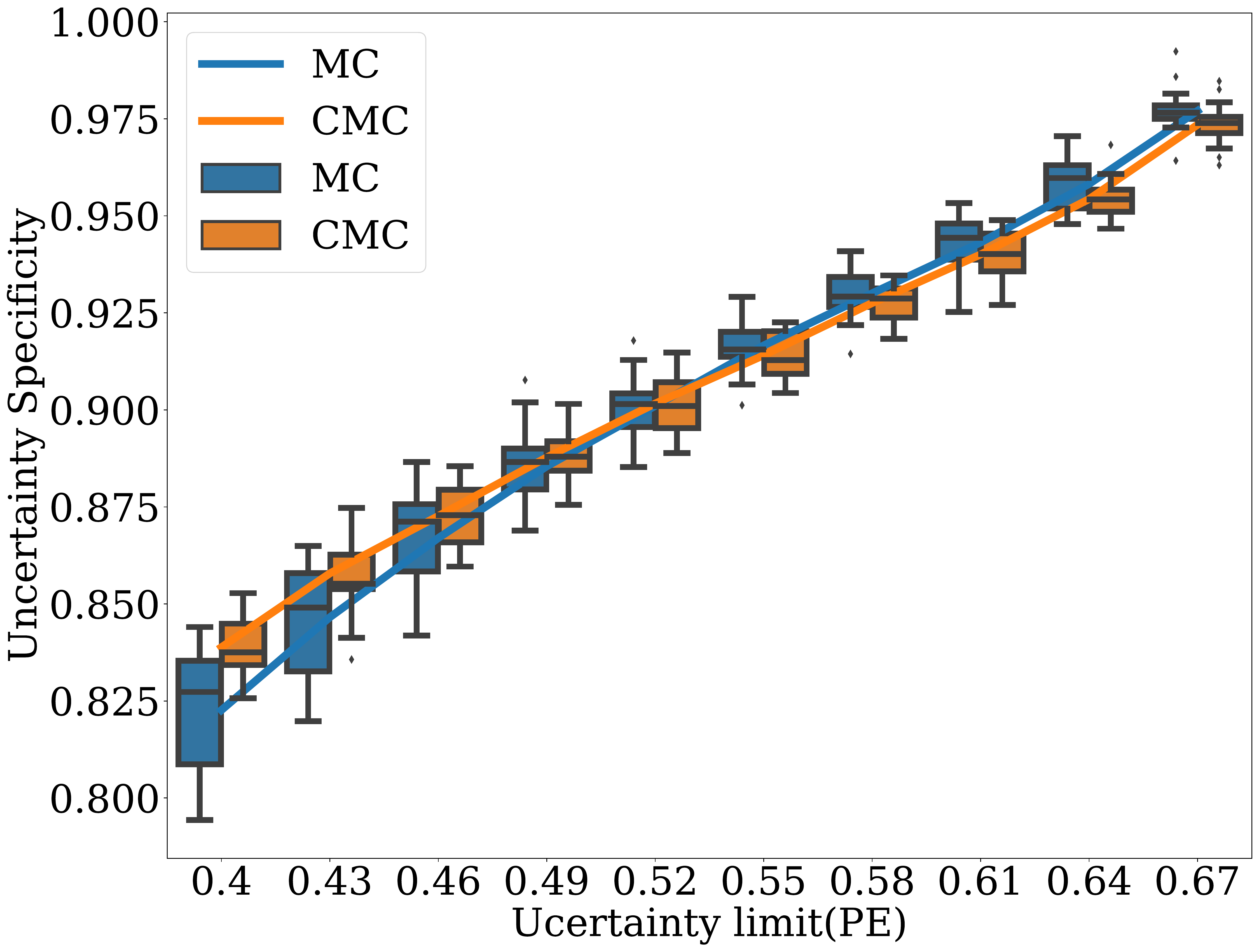}%
\label{moon_spec_uncer}}
\caption{Comparisons of uncertainty metrics between MC dropout and CMC dropout model on moon dataset. Here, the x-axis indicates the Predictive Entropy (PE) as the uncertainty threshold.}
\label{moon_uncer_result}
\end{figure*}
\begin{figure*}[!t]
\centering
\subfloat[]{\includegraphics[width=0.45\columnwidth]{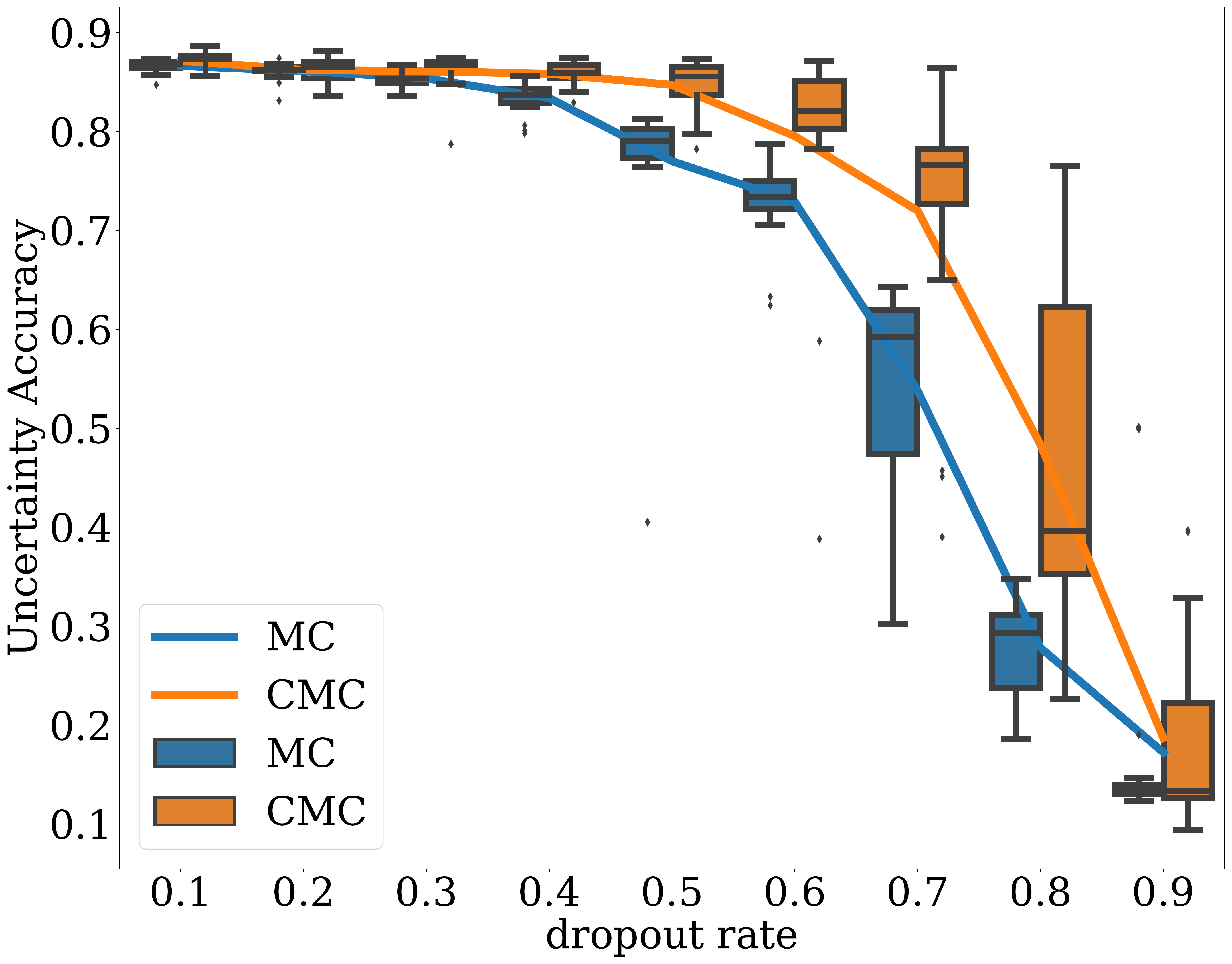}%
\label{moon_acc_drop}}
\hfil
\subfloat[]{\includegraphics[width=0.45\columnwidth]{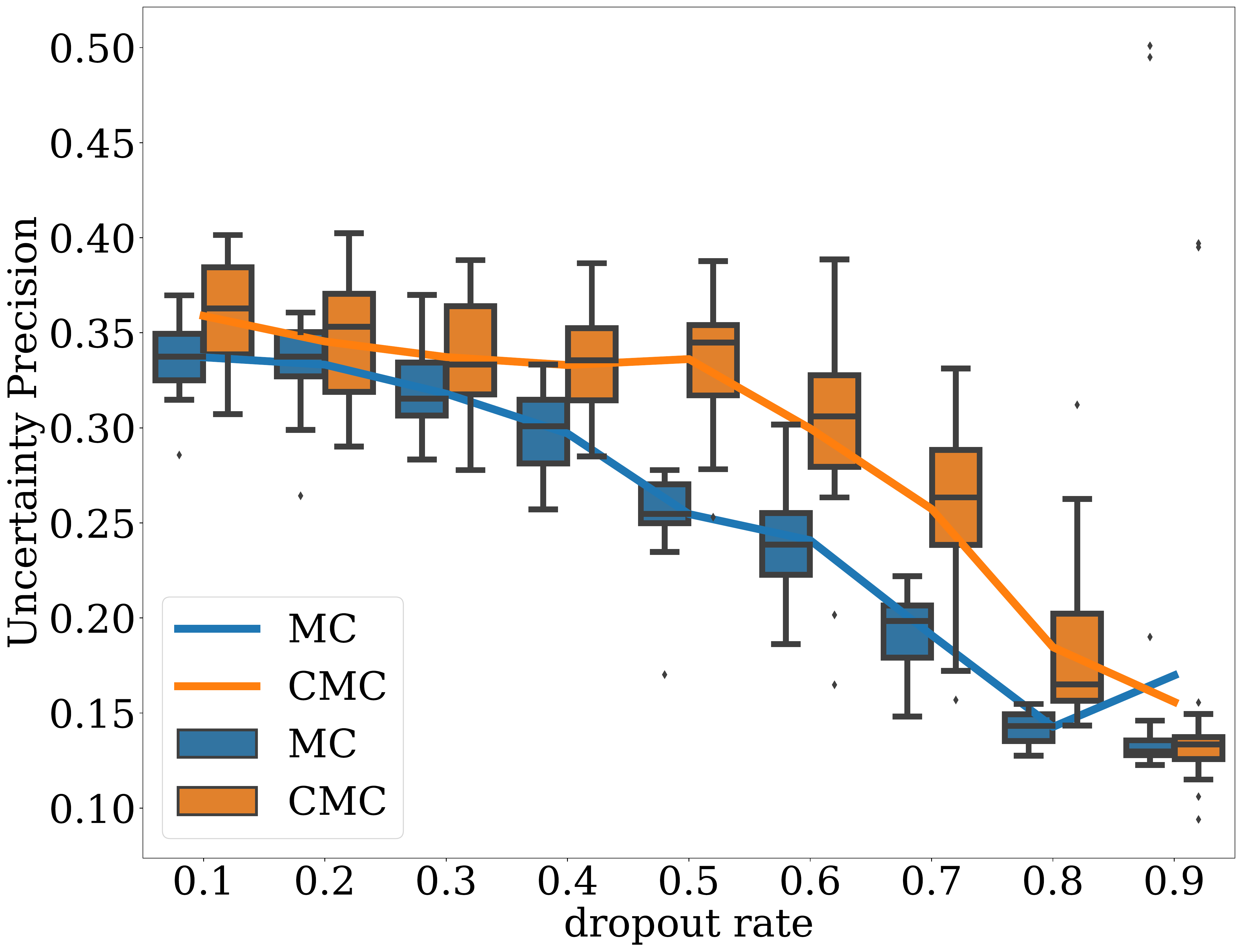}%
\label{moon_prec_drop}}
\hfil
\subfloat[]{\includegraphics[width=0.45\columnwidth]{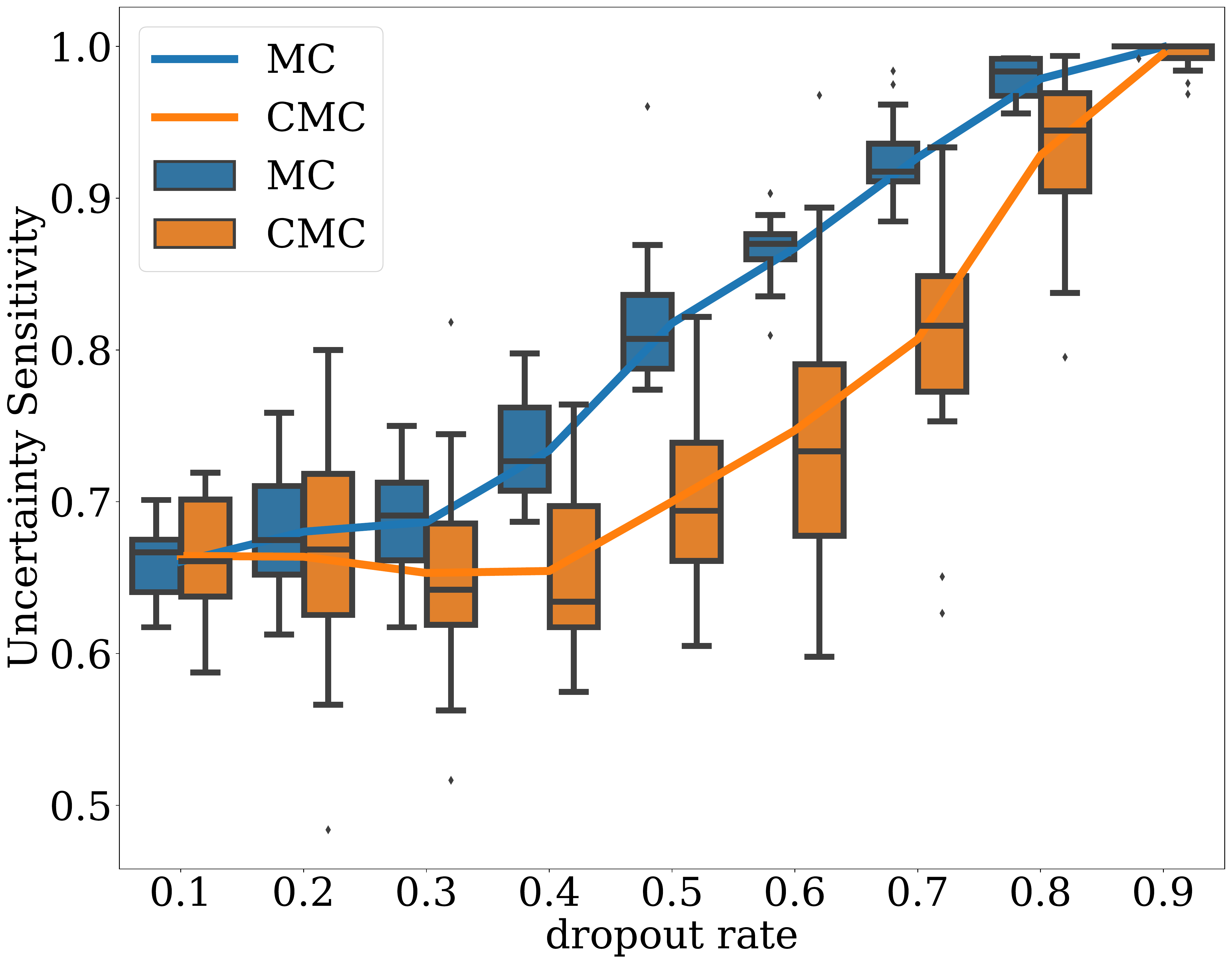}%
\label{moon_sen_drop}}
\hfil
\subfloat[]{\includegraphics[width=0.45\columnwidth]{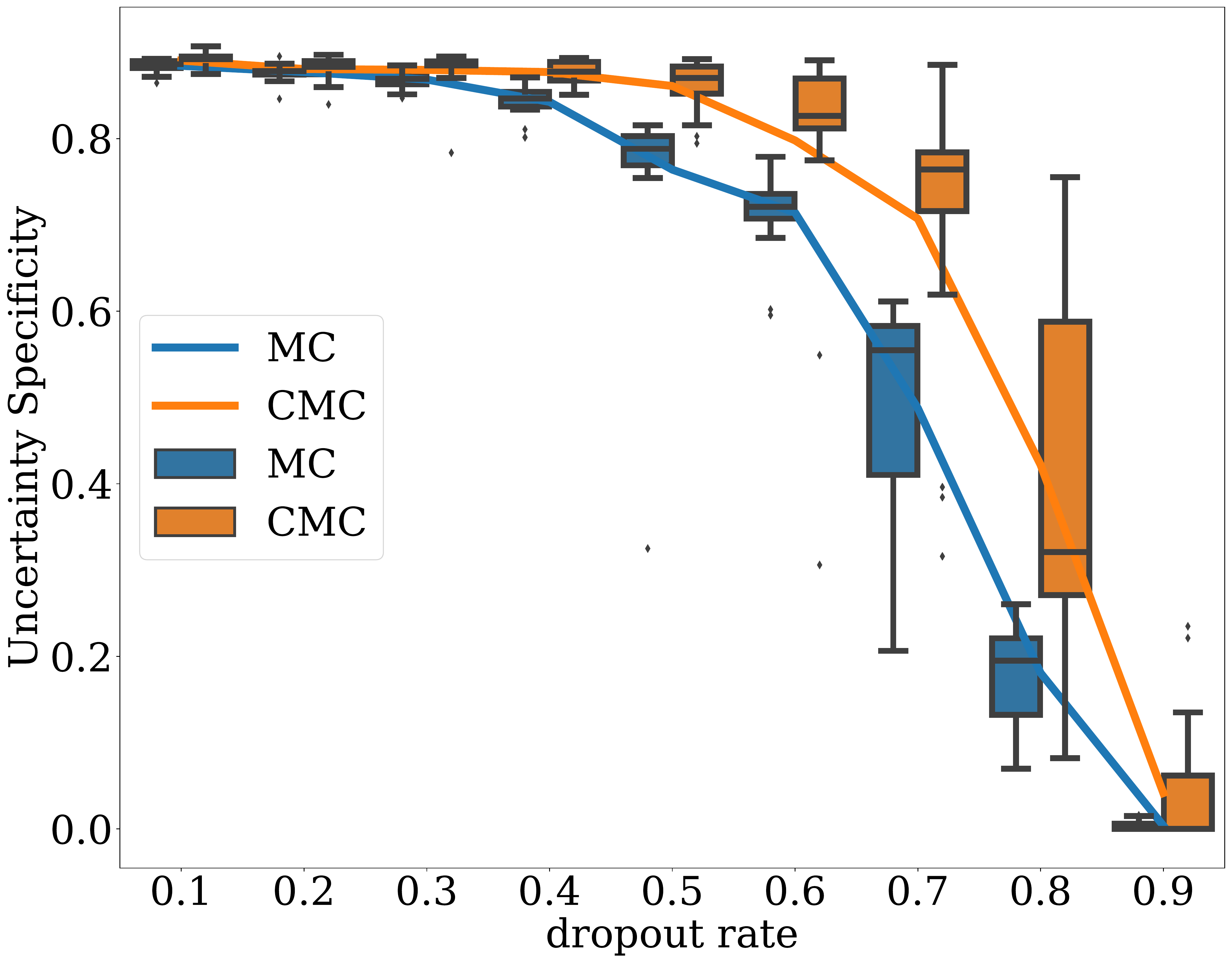}%
\label{moon_spec_drop}}
\caption{Comparisons of uncertainty metrics between MC dropout and CMC dropout on moon dataset.}
\label{moon_drop_result}
\end{figure*}

Two types of experiments are executed: fixing the dropout rate and varying the uncertainty threshold in one experiment and fixing the uncertainty threshold and varying the dropout rate in another. For the first experiment, we choose 0.3 as the dropout rate for both models. The results for 20 runs for both models are summarized in Fig. \ref{moon_uncer_result} and \ref{moon_drop_result} for the moon dataset. We also did a similar experiment, fixing the dropout rate and varying uncertainty threshold, on circle dataset with the same model and same protocol except for the noise level which was set to 0.07 in this case. Results are shown in Fig. \ref{circle_uncer_result}.

\subsection{MNIST dataset}
MNIST\cite{deng2012mnist} dataset is utilized to inspect the behavior of our process in a practical application. Here, we clearly highlight that our goal is not to achieve state-of-the-art classification performance for this dataset. Rather, we wish to compare a model with the proposed dropout process to a regular dropout layer. Consequently, the choice of model structure and different hyper-parameters is not based on any optimization strategy. LeNet-5\cite{lecun1998gradient} structure is used with the input feature of the two linear layers being 320 and 50, respectively. For our dropout layers, we constrain the number of dropout structures to 20. Dropout probability of 0.5 is used on those linear layers for both models. We implement all experiments in the Pytorch framework likewise. Negative log-likelihood is used as the loss criterion, and stochastic gradient descent is utilized as the optimization process with a learning rate of 0.001 and momentum of 0.9. The MNIST training set is randomly divided into training and validation sets with the sizes of 50000 and 10000 samples, respectively. We observe that our custom dropout model converges faster. Following that, we train our CMC model for 70 epochs and traditional dropout model for 100 epochs and saved the best model providing the minimum loss on the validation dataset.

During the test time, 100 samples are taken for each input to calculate uncertainty metrics (e.g., uncertainty accuracy, uncertainty sensitivity, uncertainty specificity, and uncertainty precision). Both models are trained, validated, and tested 20 times. Results are shown on Fig. \ref{mnist_result}. 

\section{result and discussion}
\label{sec:result}
\subsection{Two Moon and Circle Dataset}
At first, we will focus on the results for the two-moon dataset. From Fig.  \ref{moon_acc_uncer} and \ref{moon_spec_uncer}, we notice that in terms of accuracy and specificity, the CMC model presents better performance in the lower uncertainty threshold region and then follows the MC model curves closely for both metrics before slightly being lower towards the end of the threshold. Inspecting Fig. \ref{moon_prec_uncer}, it becomes clear that precision, on average, is always higher for the CMC model except for a small region in the higher threshold region. Finally, Fig. \ref{moon_sen_uncer} demonstrates that the sensitivity is also better for our method except for the lower uncertainty threshold edge, where the average value of sensitivity is slightly lower for the CMC model.

Fig. \ref{moon_drop_result} demonstrates the results of varying dropout rates from our second experiment on the moon data. We kept the uncertainty threshold fixed at 0.5. Here, we want to emphasize that at higher dropout rates, each sample of the dropout models during the training phase might not be complex enough to learn the decision boundary properly, as the model size is comparatively small. Keeping this in mind, we can observe from fig. \ref{moon_acc_drop} and \ref{moon_spec_drop} that the accuracy and specificity are similar at lower dropout rates for both models. For dropout rates greater than 0.3, both metrics become higher for the CMC model. According to Fig. \ref{moon_prec_drop}, the CMC model offers higher Precision by a good margin in most cases. For sensitivity in Fig. \ref{moon_sen_drop}, the CMC model performs worse than the MC model in most cases, but it is still comparable for the lower dropout rates.

\begin{figure*}[!t]
\centering
\subfloat[]{\includegraphics[width=0.45\columnwidth]{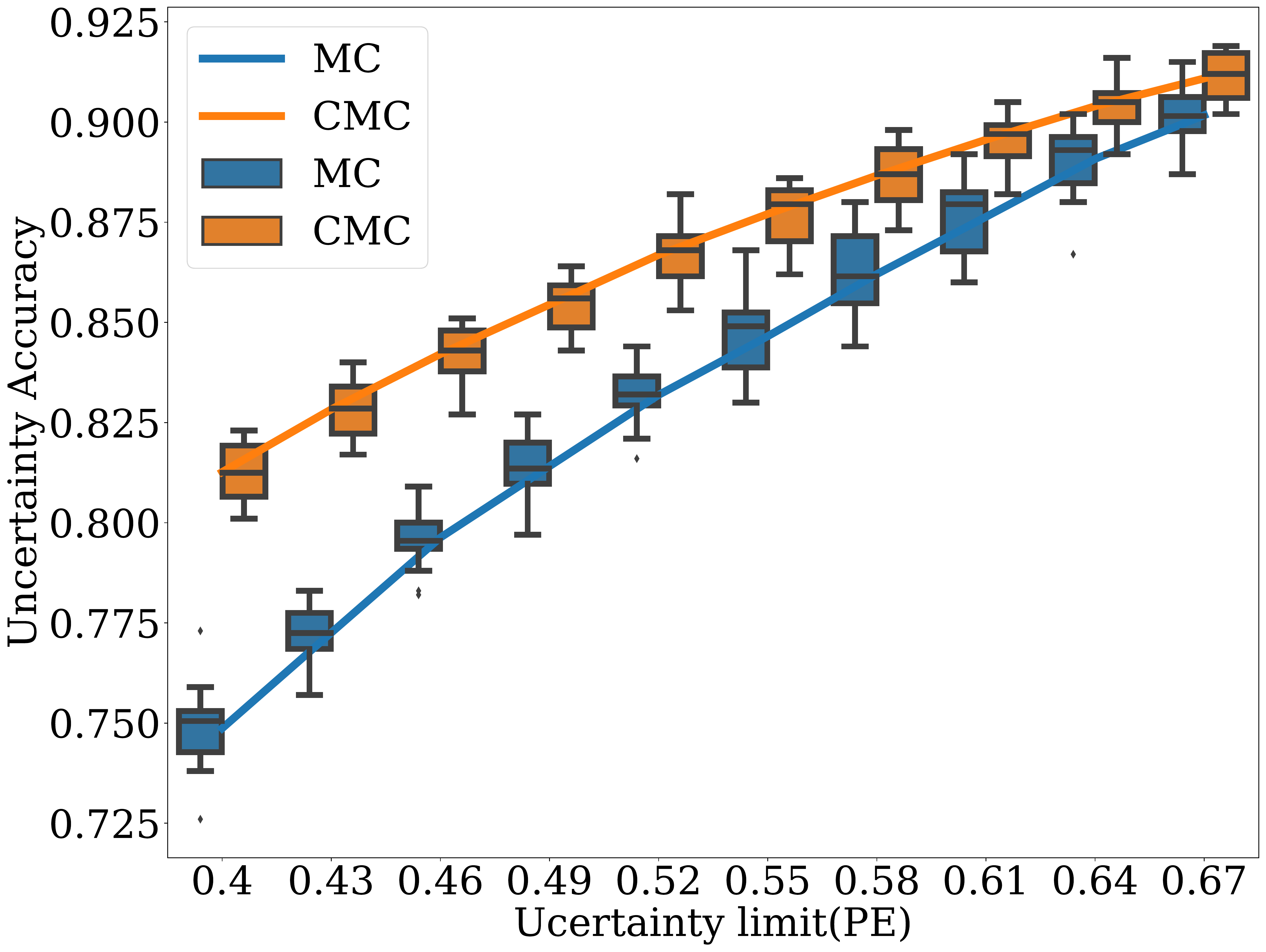}%
\label{circle_acc_uncer}}
\hfil
\subfloat[]{\includegraphics[width=0.45\columnwidth]{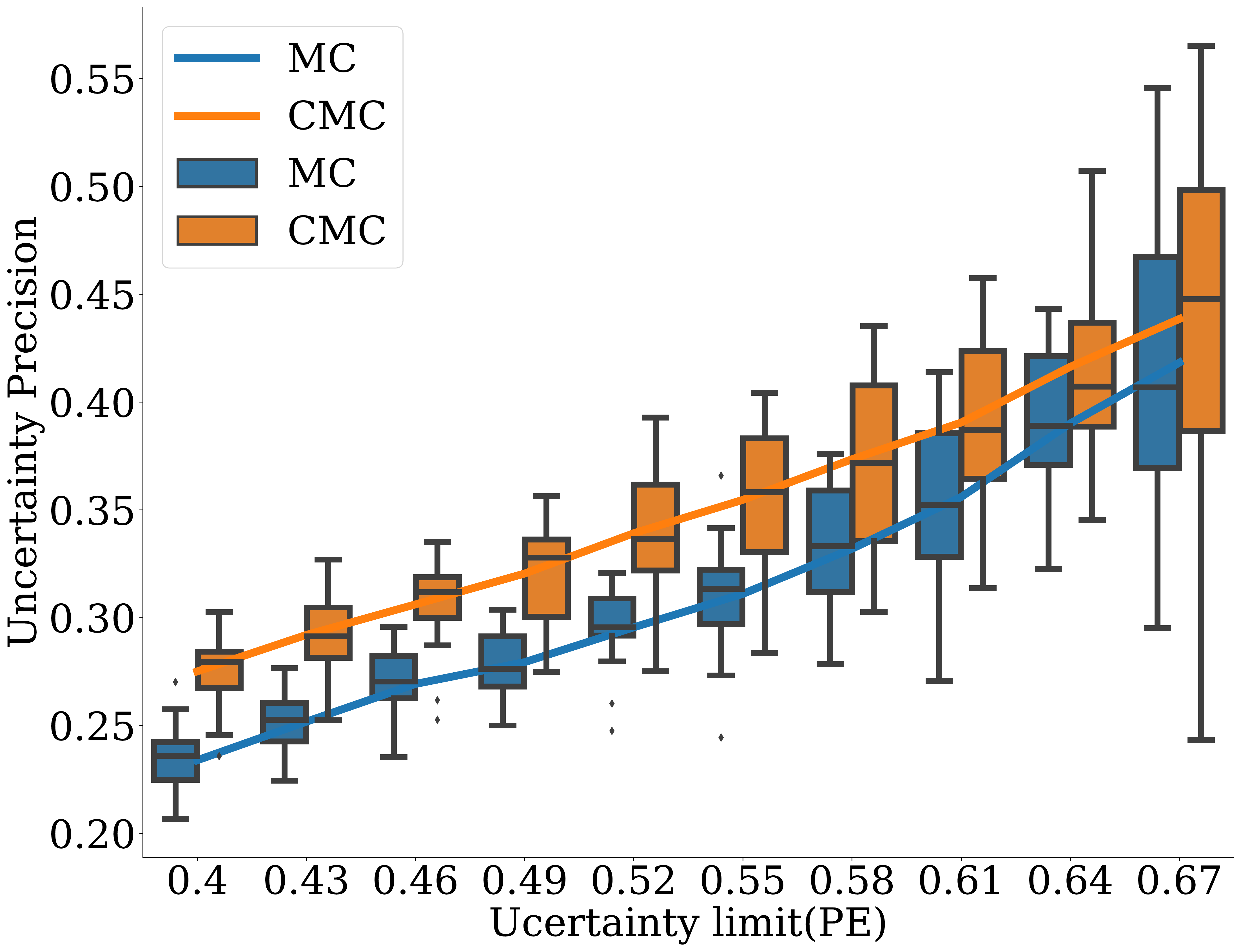}%
\label{circle_prec_uncer}}
\hfil
\subfloat[]{\includegraphics[width=0.45\columnwidth]{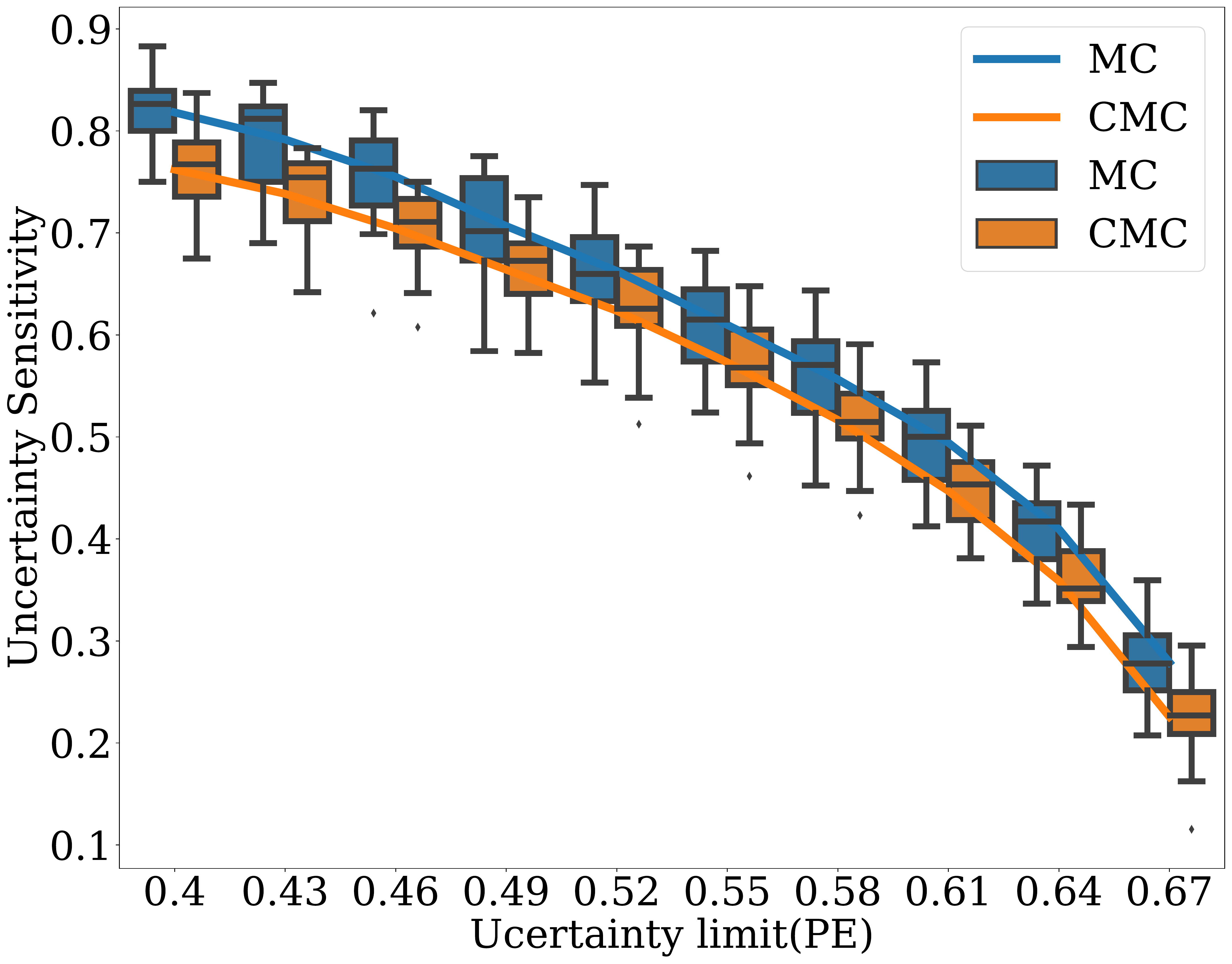}%
\label{circle_sen_uncer}}
\hfil
\subfloat[]{\includegraphics[width=0.45\columnwidth]{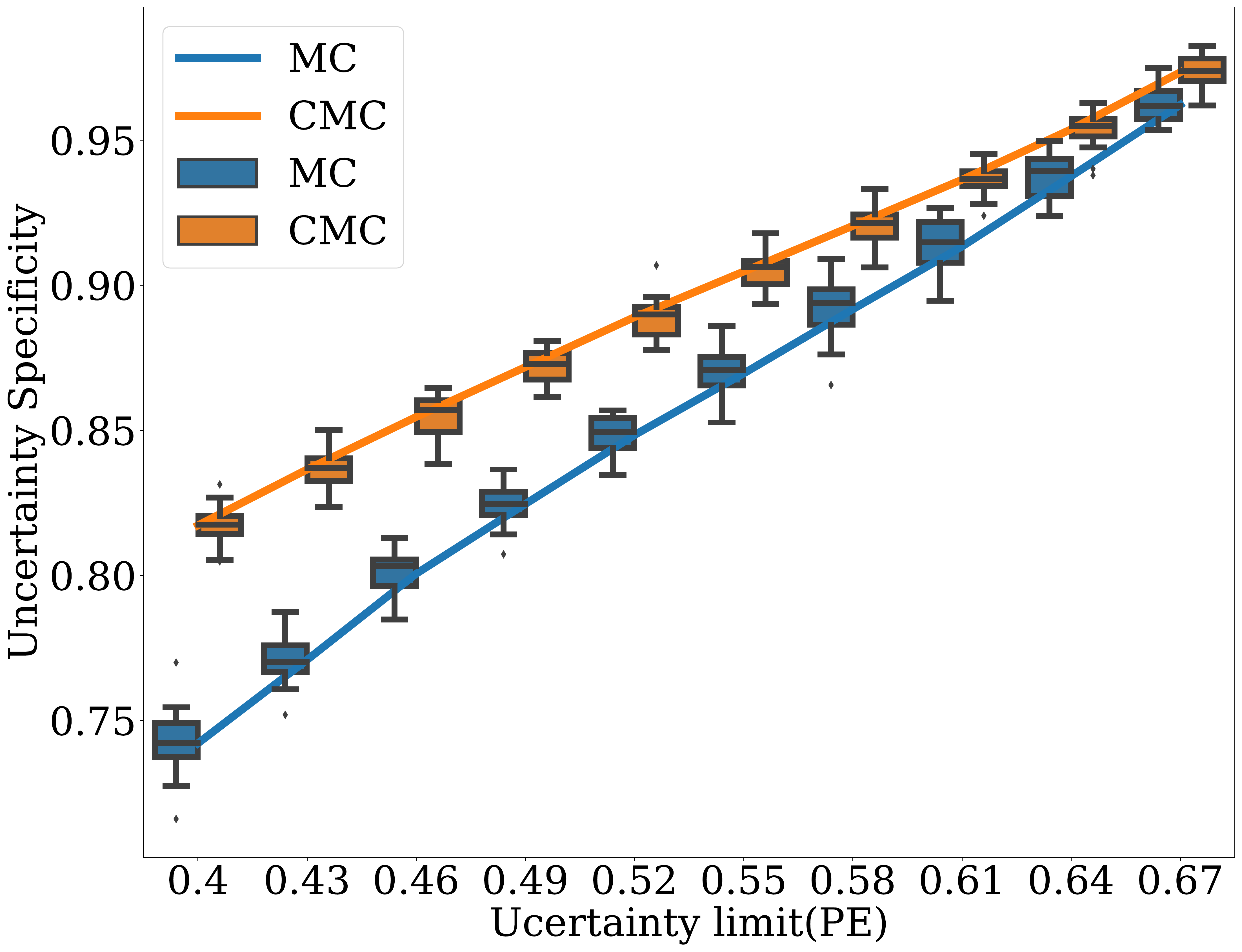}%
\label{circle_spec_uncer}}
\caption{Comparisons of uncertainty metrics between MC dropout and our CMC dropout on circle dataset. Here, the x-axis indicates the Predictive Entropy (PE) as the uncertainty threshold.}
\label{circle_uncer_result}
\end{figure*}

Studying results for the circle dataset from Fig. \ref{circle_acc_uncer} and \ref{circle_spec_uncer}, we notice that both accuracy and precision are always higher for our model by a big margin though the gap gets slightly lower towards the end of the threshold axis. Precision in Fig. \ref{circle_prec_uncer} also shows better results for the CMC model though the variation from different runs for both models is high, and values overlap quite often towards the end of the axis. Sensitivity does not offer a better result for the proposed method compared to the traditional method, as depicted in Fig. \ref{circle_sen_uncer}. Though it seems like sensitivity is worse for the CMC model compared to the MC model, we will show that this is not the case in the next section when explaining results for MNIST data. For the circle dataset, we exclude the other experiment as we observe that the results might not be conclusive for higher dropout rates due to the size limitation of the applied model.
\subsection{MNIST dataset}
\begin{figure*}[!t]
\centering
\subfloat[]{\includegraphics[width=0.45\columnwidth]{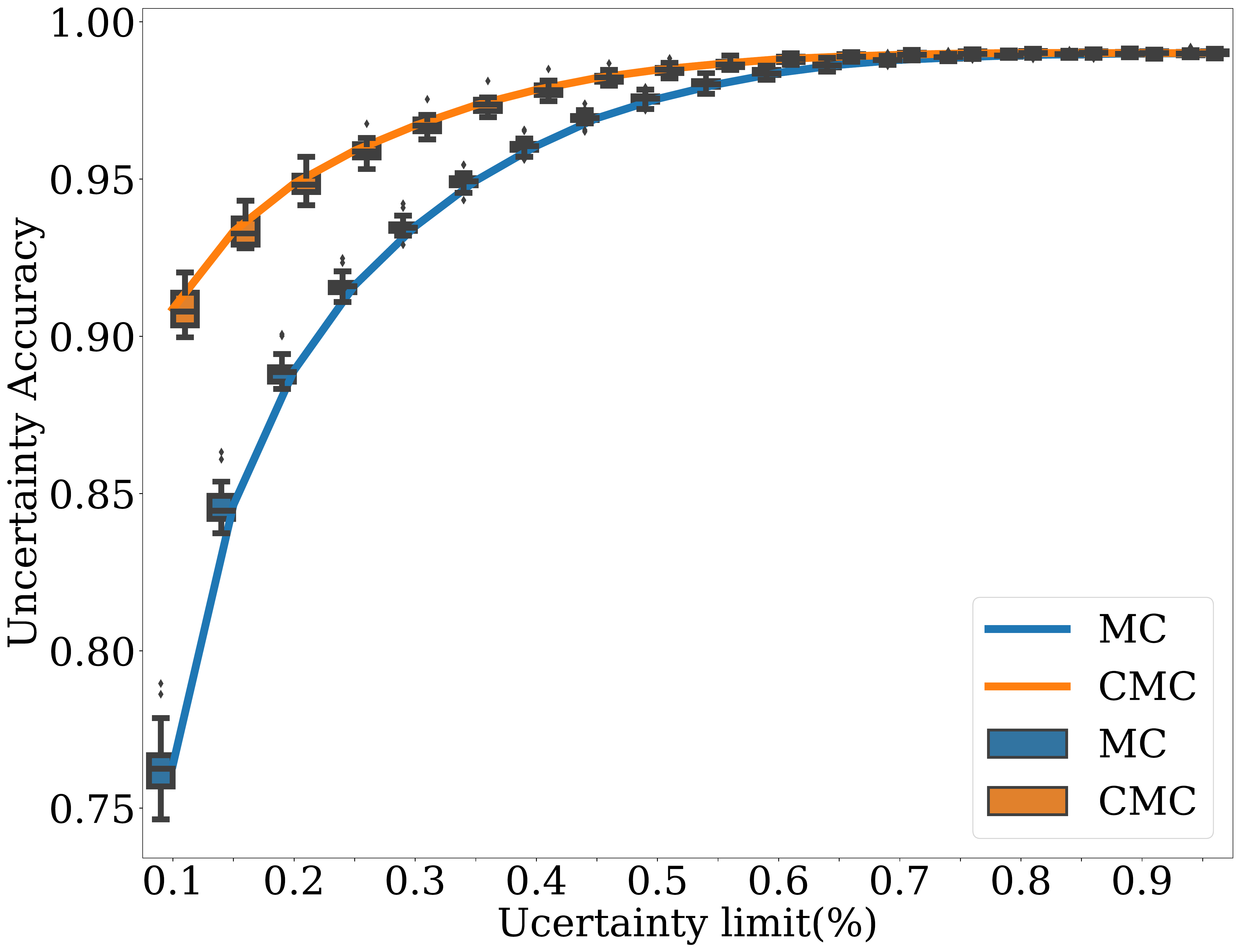}%
\label{mnist_acc}}
\hfil
\subfloat[]{\includegraphics[width=0.45\columnwidth]{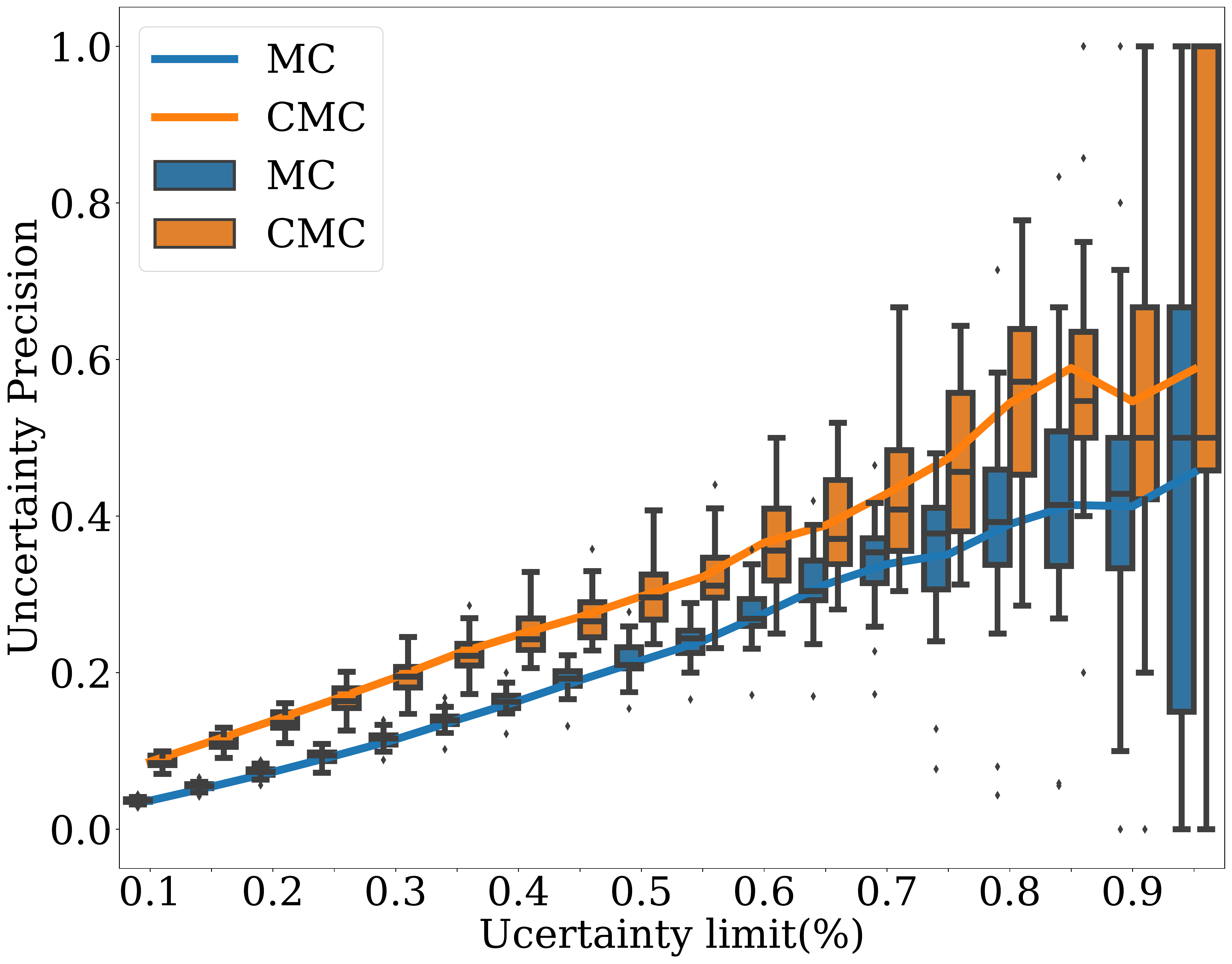}%
\label{mnist_prec}}
\hfil
\subfloat[]{\includegraphics[width=0.45\columnwidth]{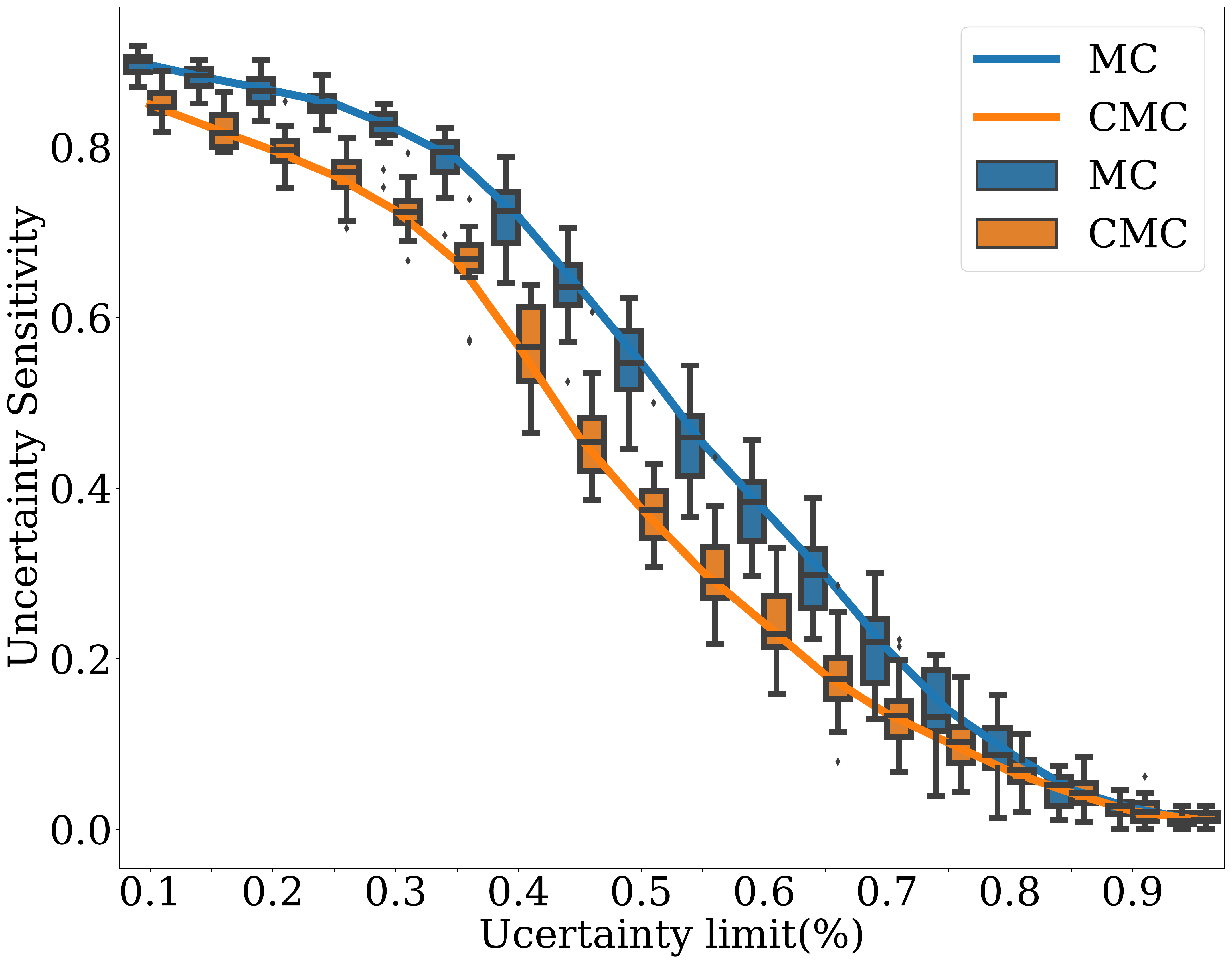}%
\label{mnist_sen}}
\hfil
\subfloat[]{\includegraphics[width=0.45\columnwidth]{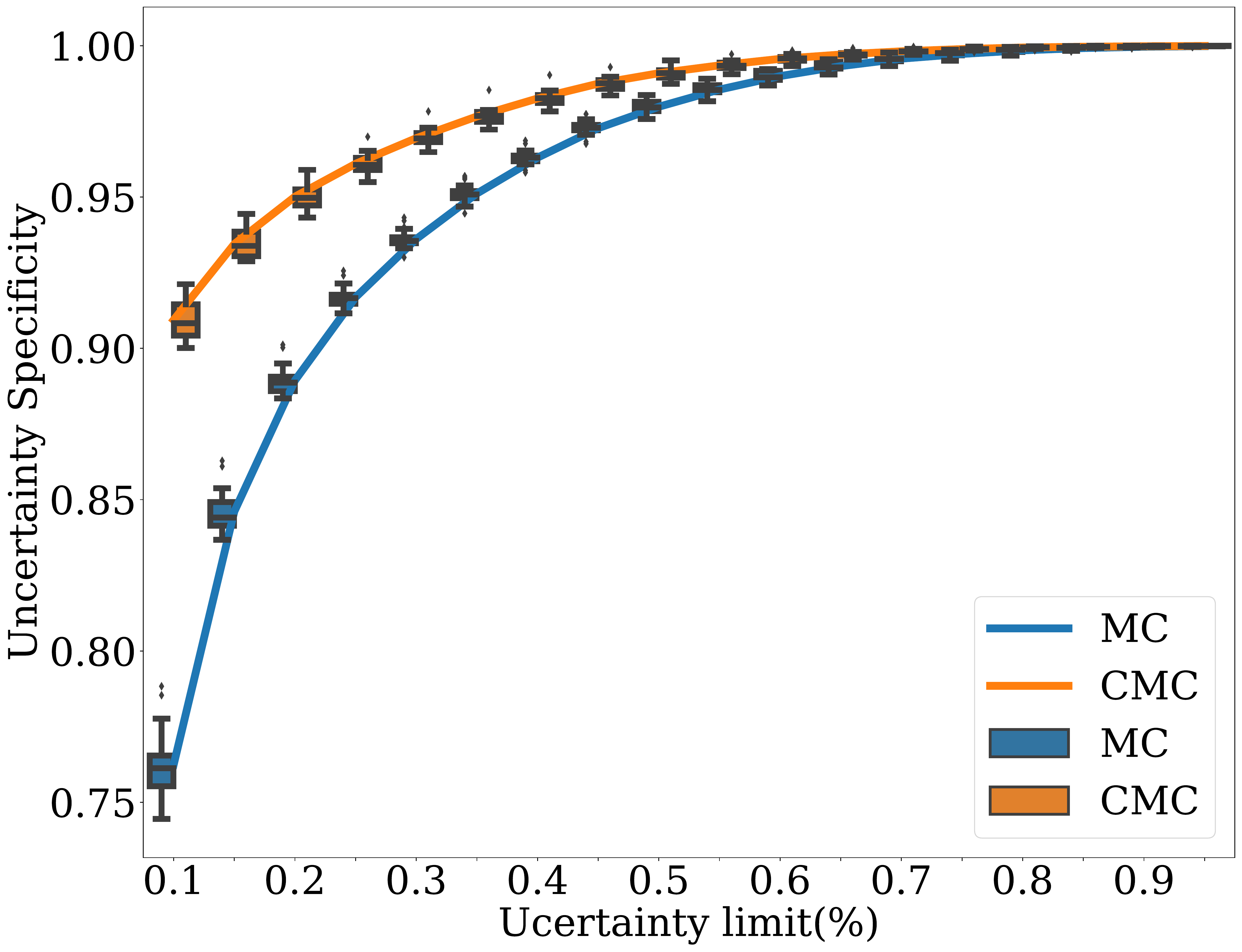}%
\label{mnist_spec}}
\caption{Comparisons of uncertainty metrics between the MC dropout and our CMC dropout on MNIST dataset. The X-axis of the figures is the percentage of the predictive entropy range that was used as the uncertainty threshold to differentiate between decided and undecided cases.}
\label{mnist_result}
\end{figure*}

 Fig. \ref{mnist_acc} and \ref{mnist_spec} depict that the accuracy and precision are much higher for CMC model in lower threshold regions. At higher thresholds, both models perform similarly in terms of these metrics. In terms of precision, the CMC model turns out to be always better though the variation from different runs is higher for bigger thresholds, and values for both models overlap. Lastly, as depicted in Fig. \ref{mnist_sen}, our model does not offer better results for sensitivity compared to the traditional method, although the gap is pretty small in most cases.  

\begin{figure}[!t]
\centering
\includegraphics[width=2.5in]{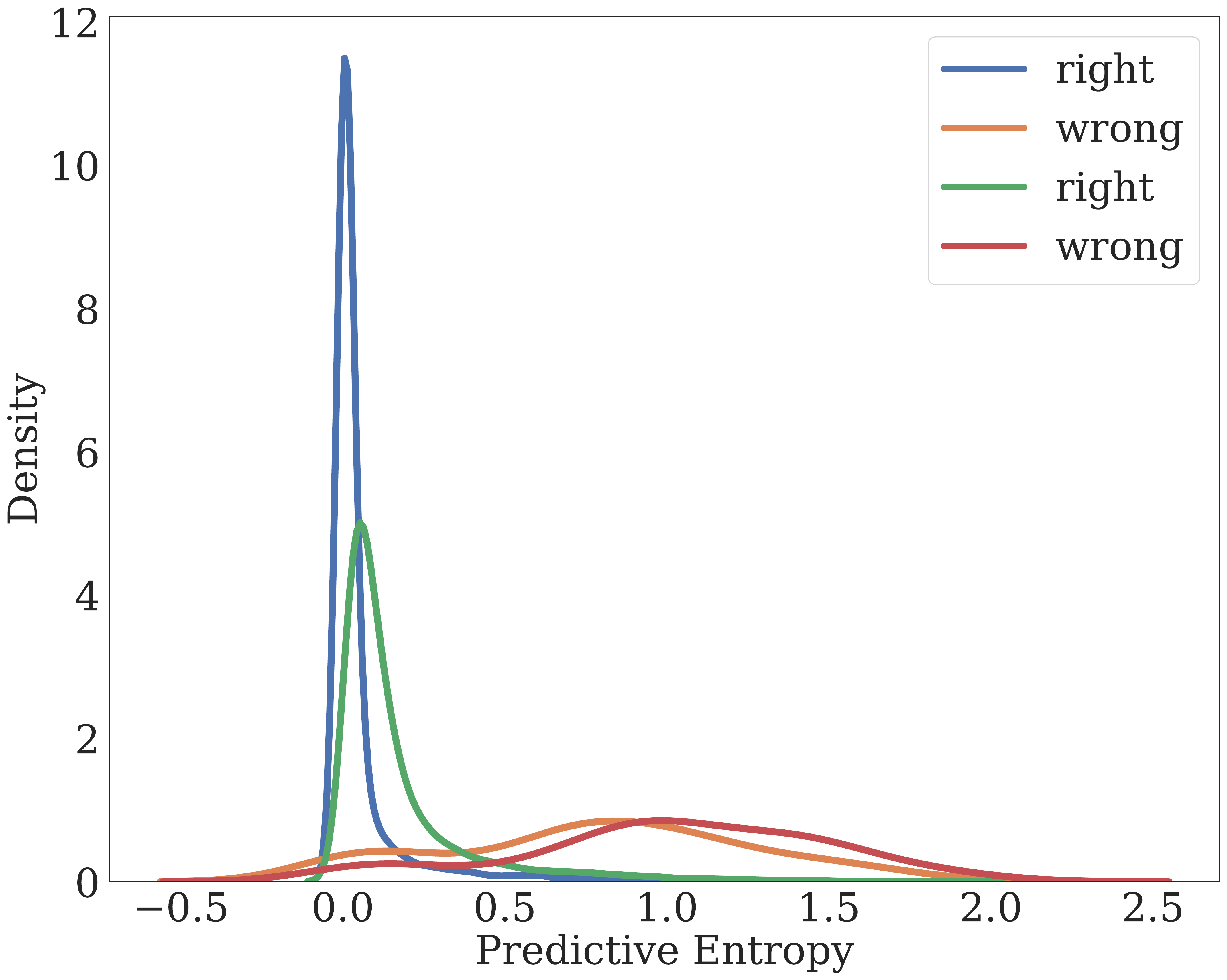}
\caption{Distributions of predictive entropy of right and wrong output. The first right-wrong group in the legend is for the CMC model, and the second one is for the MC model.}
\label{mnist_distribution}
\end{figure}

To grasp the results in Fig. \ref{mnist_result}, we need to know the distributions of predictive entropy of correct and incorrect predictions for both models based on which all the uncertainty metrics are generated. At first, we shall focus on the sensitivity metric. From distributions in Fig. \ref{mnist_distribution}, we can observe that the density of incorrect outputs for the CMC model (indicated by the first right-wrong group in the legend) is higher in the low predictive entropy region than in the MC model. This suggests that the CMC model generates the incorrect prediction with higher confidence compared to the MC model for different samples of the same input. We assume this to be the reason behind the lower sensitivity for the CMC model. Better accuracy, precision, and specificity can be explained by observing the fact that the density of correctly decided predictions for the CMC model is more concentrated in the lower predictive entropy region than in the MC model. At a higher uncertainty limit, accuracy and specificity become almost similar. This happens since there is very low or no density of right output for both models in the high predictive entropy region in Fig. \ref{mnist_distribution}.
\begin{figure}[!t]
\centering
\includegraphics[width=2.5in]{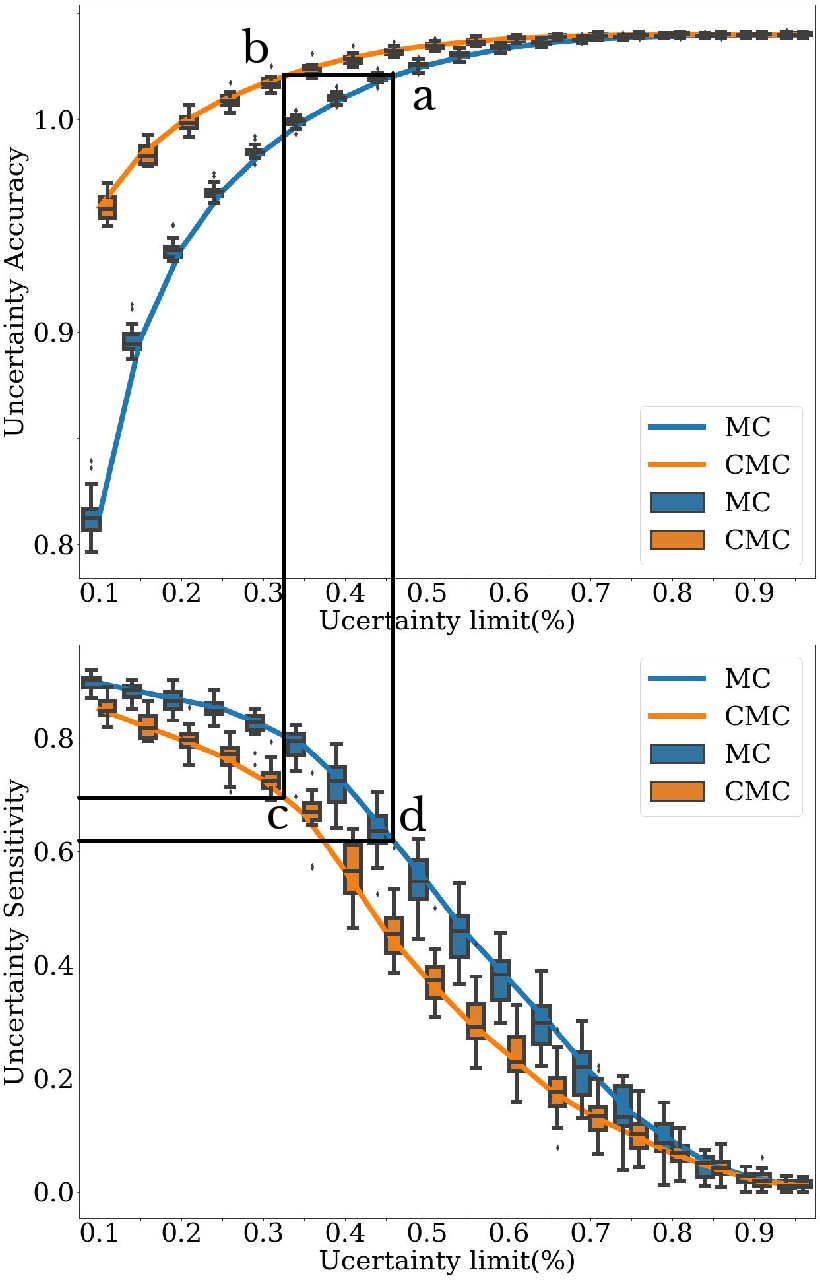}
\caption{Comparison between sensitivity and accuracy for the MNIST dataset}
\label{acc_sen_compare}
\end{figure}

At this point, it is clear that our method provides better results for all the uncertainty metrics except for sensitivity. We are well aware of the fact that sensitivity is one of the most important metrics that we expect to have a higher value for a reliable model. To show the comparison between sensitivity and accuracy of both models, we refer to Fig. \ref{acc_sen_compare} where the same accuracy level for the MC model and CMC model is indicated by points a and b, respectively. It can easily be seen that the sensitivity of the CMC model marked by point c is slightly better than the corresponding value of the MC model in point d while maintaining the same accuracy level. But, the uncertainty limit changes when we move from one model to another.

The results discussed above clearly show that the MC method using the controlled dropout layer performs better in terms of uncertainty evaluation metrics for most cases. Sensitivity did seem to be worse, but we demonstrated in Fig. \ref{acc_sen_compare} that we might gain even better sensitivity for our method just by shifting the uncertainty threshold while attaining the same accuracy as the traditional dropout method.

As an interesting point, we can store different numbers of configurations for different layers within an NN using the proposed approach. The numbers of stored configurations in all our experiments are somewhat arbitrary. Choosing the optimal number of structures or analyzing the effect of the number of stored configurations on the uncertainty metrics is beyond the scope of this study. At this stage, we consider this as one of the regular hyper-parameters that needs to be manually set when defining the model.
\section{Conclusion}
\label{sec:Conclusion}
Uncertainty quantification using MC dropout has gained much attention over the last few years. In this work, we introduced the controlled dropout layer with customized and fixed dropout configurations for each dropout layer. The algorithm to generate the proposed dropout layer has been presented. Performance analysis using uncertainty evaluation metrics on synthetic and realistic datasets confirms that the proposed method has the potential to offer better results compared to the traditional uncertainty quantification techniques.

In the future, we wish to investigate the effect of our dropout layer on regression cases. Moreover, studying the overfitting reduction capability of the dropout method is affected when using the proposed layer is important. The choice of an optimized number of configurations for each dropout layer also needs to be investigated in the future.


\bibliographystyle{IEEEtran}
\bibliography{citation}

\vfill

\end{document}